\documentclass[runningheads]{llncs}

\usepackage[T1]{fontenc}
\usepackage{amsmath}
\usepackage{hyperref}
\usepackage{subcaption}
\usepackage{multirow}

\def\doi#1{\href{https://doi.org/\detokenize{#1}}{\url{https://doi.org/\detokenize{#1}}}}
\usepackage{graphicx}
\usepackage{listings}
\begin{document}
\title{A Machine Learning and Computer Vision Approach to Geomagnetic Storm Forecasting}
%
%
\author{Kyle Domico \and Ryan Sheatsley \and \\ Yohan Beugin \and Quinn Burke \and Patrick McDaniel}
\authorrunning{K. Domico et al.}
\titlerunning{Geomagnetic Storm Forecasting}
%
\institute{The Pennsylvania State University, University Park, PA 16802, USA\\ 
\email{\{domico,sheatsley,ybeugin,qkb5007\}@psu.edu, mcdaniel@cse.psu.edu}}
\maketitle              
\begin{abstract}
Geomagnetic storms, disturbances of Earth's magnetosphere caused by masses of charged particles being emitted from the Sun, are an uncontrollable threat to modern technology. Notably, they have the potential to damage satellites and cause instability in power grids on Earth, among other disasters. They result from high sun activity, which are induced from cool areas on the Sun known as \emph{sunspots}. Forecasting the storms to prevent disasters requires an understanding of how and when they will occur. However, current prediction methods at the National Oceanic and Atmospheric Administration (NOAA) are limited in that they depend on expensive solar wind spacecraft and a global-scale magnetometer sensor network. In this paper, we introduce a novel machine learning and computer vision approach to accurately forecast geomagnetic storms without the need of such costly physical measurements. Our approach extracts features from images of the Sun to establish correlations between sunspots and geomagnetic storm classification and is competitive with NOAA's predictions. Indeed, our prediction achieves a 76\% storm classification accuracy. This paper serves as an existence proof that machine learning and computer vision techniques provide an effective means for augmenting and improving existing geomagnetic storm forecasting methods. 

\end{abstract}
\section{Introduction}

Geomagnetic storms are a solar weather event that occur when masses of charged particles are emitted from the Sun (often called solar flares or coronal mass ejections) and interact with the Earth's magnetic field. The effects of the storms range from inducing voltage into power grids on Earth to more catastrophic failures like causing transformers to explode or altering orbital tracks of satellites (which could lead to collisions with other debris or spacecraft~\cite{noauthor_noaa_nodate,noauthor_geomagnetic_nodate}). Forecasting geomagnetic storms is therefore crucial to ensuring proper operation of these technological systems.

Scientists at the National Oceanic and Atmospheric Administration (NOAA) predict geomagnetic storms by collecting atmospheric measurements from magnetometers at several stations across the globe. Additionally, a real-time solar wind network of spacecraft collects atmospheric samples of high-energy particles emitted from the Sun. Using this information, they can forecast storms for the next 3 days~\cite{noauthor_3-day_nodate} based on a global average across all magnetometers and spacecraft measurements. However, ground-based magnetometers are aging and becoming unreliable~\cite{engebretson_future_2017}.

Based on the observation that sunspot activity is correlated with high solar activity~\cite{us_department_of_commerce_space_nodate}, we study if it is possible to use sunspot features on images of the Sun to predict geomagnetic storms. In this paper, we leverage computer vision on active sunspots in images to predict geomagnetic storms. Specifically, we pair state-of-the-art supervised learning models with direct images of the Sun to predict storms, forgoing the need for a global-scale magnetometer and a solar wind spacecraft sensor network. The prediction algorithm consists of two sequential layers: an image processing layer followed by a prediction layer. The image processing layer is composed of a series of image processing algorithms to extract sunspot features. The prediction layer then uses machine learning to predict if a geomagnetic storm will occur in the next 24 hours.

To evaluate the efficacy of our approach, we used publicly available images of the Sun from NASA's Solar Dynamics Observatory (SDO) \cite{pesnell_solar_2012}. The SDO is a single satellite that collects a variety of Sun images every 15 minutes \cite{zell_sdo_2015}. With 2843 images of the Sun, spanning from January 2012 to April 2021, our models reached an overall accuracy of 76\% across classifications. Our approach demonstrates that machine learning techniques are an effective means towards forecasting geomagnetic storms. 

In this work, we contribute the following:
\begin{enumerate}
    \item We show that active sunspot features can be reliably identified from images of the Sun and are accurately correlated with geomagnetic storm classification.
    \item We introduce a machine learning based technique that can forecast geomagnetic storms from image data of just a single satellite. 
    \item We demonstrate that machine learning techniques are an effective means for geomagnetic storm forecasting through a comprehensive evaluation.
\end{enumerate}

\section{Background}

\begin{figure}[t]
\centering
\begin{subfigure}{.5\textwidth}
  \centering
  \includegraphics[width=.9\linewidth]{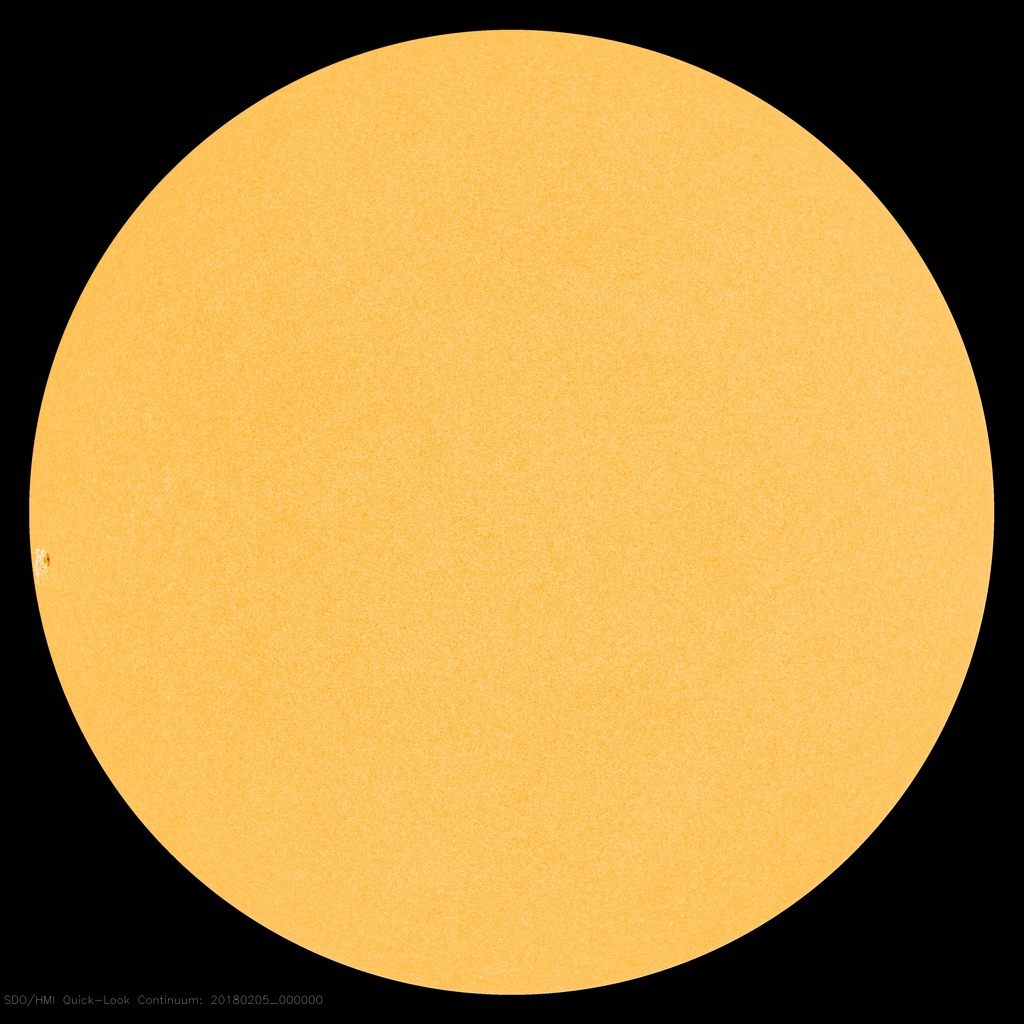}
  \caption{\textbf{No Storm Detected (02/05/2018)}}
  \label{fig:nostorm}
\end{subfigure}%
\begin{subfigure}{.5\textwidth}
  \centering
  \includegraphics[width=.9\linewidth]{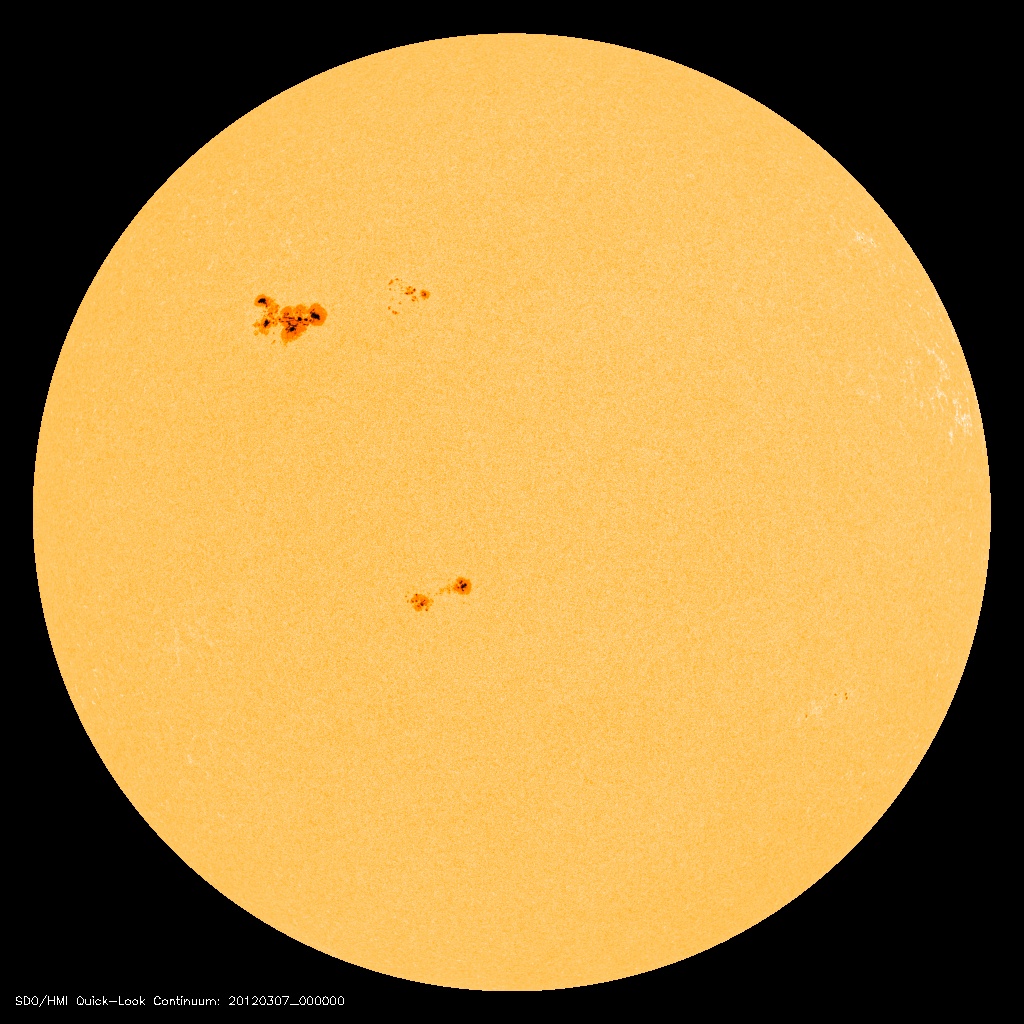}
  \caption{\textbf{Storm Detected (03/07/2012)}}
  \label{fig:storm}
\end{subfigure}
\caption{Image of the Sun from the Solar Dynamics Observatory Image (SDO) on days with different storm classifications. The difference in visible sunspots can be seen in the storm day (b) as opposed to the no storm day (a).}
\label{fig:stormVSnostorm}
\end{figure}

\subsection{Solar Weather and Prediction Methods}

Solar weather describes the time-varying conditions of space in close proximity to Earth. The conditions are influenced by activity at the Sun that spews out gas and charged particles from its surface into space---which is referred to as a \textit{solar flare}. The energy originates from \textit{sunspots} that represent cooler areas on the Sun's surface. Sunspots themselves are caused by the tangling and crossing of magnetic field lines, which produce interference from which solar flares or coronal mass ejections (CME) arise~\cite{us_department_of_commerce_space_nodate}.

Different magnitudes of solar flares exist, varying the effects observed on Earth. Environmental disturbances caused by solar flares are categorized into three events: geomagnetic storms, solar radiation storms, and radio blackouts \cite{us_department_of_commerce_space_nodate}. The Space Weather Prediction Center (SWPC) at NOAA classifies each of these events into numbered scales, analogous to how the severity of hurricanes, tornadoes, and earthquakes are measured. We focus our attention on geomagnetic storms because of their prevalence when active sunspot numbers are high \cite{us_department_of_commerce_space_nodate}, as illustrated in \autoref{fig:stormVSnostorm}. Specifically, we observe an opportunity to use the sunspots (rather, images thereof) as tool for forecasting future storms.

Geomagnetic storm magnitude is determined by the Kp-Index measured at the time of the storm. The Kp-Index quantifies the disturbance in Earth's magnetic field on a scale from 1 to 9, 9 being the strongest of disturbances. According to the SWPC, geomagnetic storms are classified as such when the measured Kp-Index is greater than or equal to 5 \cite{noauthor_noaa_nodate}. The SWPC at NOAA currently has methods to forecast the Kp-Index for the next 3 days, and issue warnings when the Kp-Index value is expected to be greater than or equal to 5.

According to the SWPC, methods to predict and estimate the Kp-Index require a collection of ground-based magnetometer measurements from stations around the world, and real-time solar wind measurements from a network of orbital spacecraft \cite{noauthor_3-day_nodate}. Magnetometers measure the Earth's magnetic field strength, and solar wind spacecraft measure the properties of solar wind at various locations around Earth's L1 orbit. Orbital spacecraft and magnetometer stations used to collect data are not only expensive but can be unreliable and become inefficient as they are aging \cite{engebretson_future_2017}.

\subsection{Computer Vision and Image Processing}

Computer vision is a sub domain within artificial intelligence that enables computers to extract meaningful information from digital inputs such as images and videos. Edge detection algorithms became the forefront of image processing because of their usefulness in object recognition. They work by calculating gradients on pixels to determine the change in pixel intensity as the distance from each pixel increases. This proved to be useful to detect edges in images. Contour mapping algorithms are also useful when an edge-detected image is provided, as these algorithms fill in and count the closed edges of an image.

\subsection{Machine Learning}

Supervised learning aims to make predictions off of data that has truth labels. In this way, an algorithm controls weighted parameters corresponding to features of the dataset. These weights are adjusted to guide predictions based off of the truth labels matching each data point, hence the name of supervised learning. In our study, we use \emph{Support Vector Machines (SVMs)} to formulate predictions. The goal of SVMs is to create a n-dimensional hyperplane equation \(\phi\) of tunable weights \(\theta\) and bias \(b\) such that the distance, or margin, \(d\) is defined as:
\begin{equation}
    \phi(x_i) = \theta^T x_i + b
\end{equation}
\begin{equation}
    d(\phi(x_i)) = \frac{|\theta^T x_i + b|}{||\theta||_2}
\end{equation}

Where \(x_i\) is the i-th sample of the dataset and \(||\theta||_2\) denotes the Euclidean norm of the weight vector \(\theta\). From these equations, the SVM iterates to find the optimal weights \(\theta^*\) to maximize the minimum distance between samples \cite{suthaharan_support_2016}:
\begin{equation}
    \theta^* = \arg \max_\theta [\arg \min_i d(\phi(x_i)) ]
\end{equation}

Unsupervised learning differs from supervised learning in that there are no truth labels, and the learner must find some hidden structure among data features to make sense of it.

\section{Methodology}

\begin{figure}[t]
    \centering
    \includegraphics[width=\textwidth,height=\textheight,keepaspectratio]{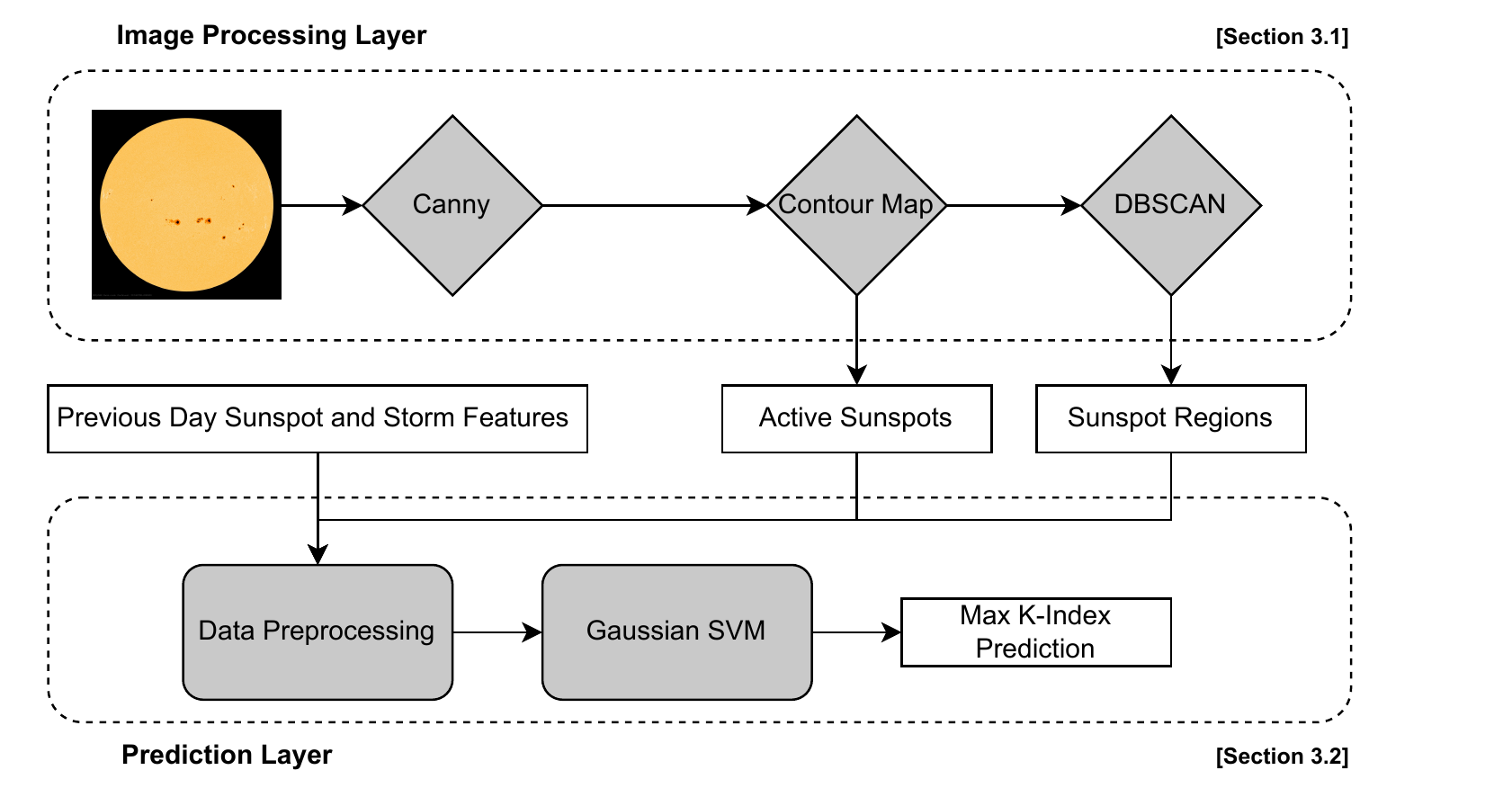}
    \caption{Our approach to forecasting geomagnetic storms leverages a 2-layer prediction pipeline and uses images of the Sun taken by the Solar Dynamics Observatory (SDO).}
    \label{fig:pipeline}
\end{figure}

Our approach consists of two layers (see \autoref{fig:pipeline}): the Feature Extraction Layer and the Prediction Layer. In the first layer, we leverage image processing and unsupervised learning algorithms to extract sunspot features from an image of the Sun. Next, we use a supervised learning algorithm to exploit these features and learn their correlation with Kp-Indices to predict geomagnetic storms.

To make the prediction, we use the sunspot features of the previous and present day to take into account the evolution of the Sun's activity:
\begin{enumerate}
    \item Previous Day Active Sunspots
    \item Previous Day Active Sunspot Regions
    \item Previous Day Storm Existence
    \item Present Day Active Sunspots
    \item Present Day Active Sunspot Regions
\end{enumerate}

\subsection{Image Processing Layer}

To extract the needed sunspots features from the image of the Sun, we must first estimate their boundaries, determine their number, and cluster them into sunspot regions. We are interested in the number of sunspots because it is believed that there is a correlation between their appearance and solar activity \cite{us_department_of_commerce_space_nodate}. Additionally, clustering sunspots is important to determine the number of active sunspot regions. More active regions on the Sun indicate a higher probability of a solar flare to be produced \cite{noauthor_solar_nodate}.

\subsubsection{Edge Detection}

To locate the sunspots on the image, we use the Canny Edge Detection (CED) algorithm. We use this algorithm because images of the Sun taken by the Solar Dynamics Observatory (SDO) contain noise from white and inactive sunspot regions that we do not want to count towards the total active sunspot count. CED was designed to mitigate the influence of such inactive regions; the algorithm first applies noise reduction via a Sobel Kernel and then finds the edge gradient based on the function \(P\), defined as the pixel intensity values at position \((x,y)\) on the image. The gradient \(G(P)\) and direction \(\theta\) is then computed by:
\begin{equation}
    G(P)=\sqrt{\left(\frac{\partial P}{\partial x}\right)^2+\left(\frac{\partial P}{\partial y}\right)^2}
\end{equation}
\begin{equation}
    \theta=\tan^{-1}\left(\frac{\left(\frac{\partial P}{\partial y}\right)}{\left(\frac{\partial P}{\partial x}\right)}\right)
\end{equation}

Once the edge gradient and angles are computed, the algorithm set to zero the pixels that are not local maxima of the \(G\) function, a method defined as non-maximum suppression. The result is a binary image with thin edges, but the algorithm then uses hysteresis thresholding to find edges that were not detected previously. This process recalls the gradient function \(G\) and requires a minimum value input. Since our interest is in active sunspot regions (represented by darker spots), we specify this minimum value to be 300 as we observed that inactive region borders have an edge-gradient value just above 200. This segments the clear dark sunspots seen in the Outlined Sun Image in \autoref{fig:iplayer}, since edge-gradient values at inactive sunspot borders will now not be recognized as edges.

\subsubsection{Topological Structure Analysis}

\begin{figure}[t]
    \centering
    \includegraphics[width=\textwidth,height=\textheight,keepaspectratio]{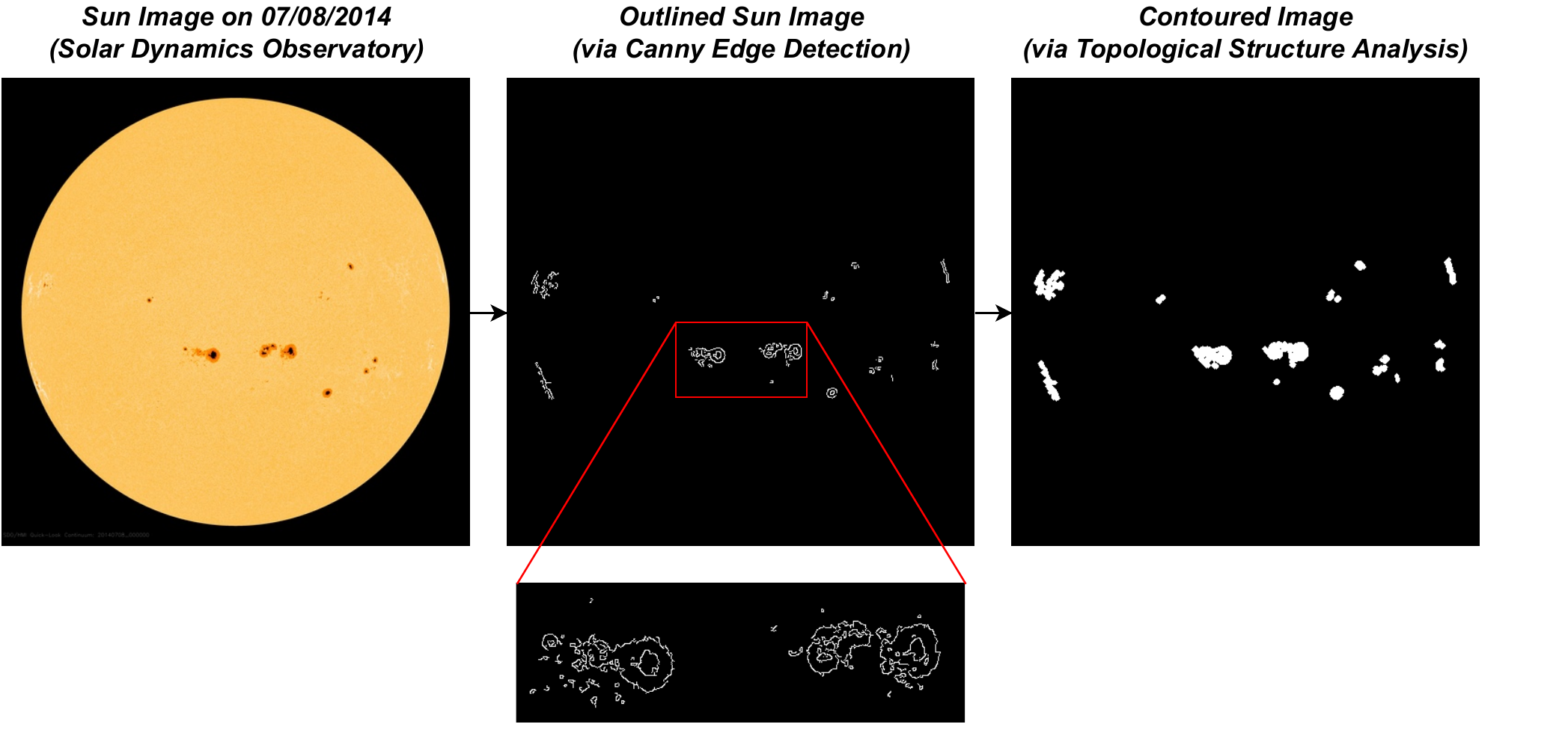}
    \caption{Step by step image representation of the image processing algorithms used to extract active sunspot features.}
    \label{fig:iplayer}
\end{figure}

The edge detection algorithm produces a binary image where the sunspots are outlined. As the magnitude of solar activity is correlated with the number of sunspots \cite{us_department_of_commerce_space_nodate}, we now want to determine the number of active sunspots. Since they appear in dark contours \cite{noauthor_solar_nodate}, we use the \textit{topological structure analysis for binary images} algorithm designed for its ability to count the number of contours in images \cite{suzuki_topological_1985}. The algorithm produces another binary image with quantified, outlined topological structures (contours) seen in the Contoured Image in \autoref{fig:iplayer}. This lets us extract the number of sunspots to be fed into the prediction layer.

\subsubsection{DBSCAN}

To extract the unknown number of active sunspot regions on the image, we use an unsupervised learning algorithm. The regions or clusters we will consider are the white pixels in the binary image produced from the topological structure analysis. The Density-Based Spatial Clustering of Applications with Noise (DBSCAN) algorithm provides an implementation of how to find clusters in the data without specifying the number of regions \cite{schubert_dbscan_2017}. In order to do this, a distance parameter \(\epsilon\) is passed into the algorithm. This parameter helps us define a function \(N_{\epsilon}(p)\) that determines the number of points within \(\epsilon\) distance of pixel \(p\):

\begin{equation}
    N_{\epsilon}(p) = \{q \in D | dist(p,q) \le \epsilon\}
\end{equation}

In addition, another parameter \(minPts\) is defined as the number of points within a distance \(\epsilon\) a cluster is to be considered a cluster. With this parameter, we define pixel \(p\) as \emph{density-reachable} with respect to pixel \(q\) if the following conditions are satisfied:

\begin{enumerate}
    \item \(p \in N_{\epsilon}(q)\)
    \item \(|N_{\epsilon}(q)| \ge minPts\)
\end{enumerate}

Iterating through each white pixel, clusters are established and the noise created from other white pixels not part of a region of sunspots are filtered out. The algorithm produces an integer number of clusters, that we will define as the number of sunspot regions and pass as feature to the prediction layer.

\subsection{Prediction Layer}

With the number of active sunspots and active sunspot regions extracted from the image of the Sun, the next layer of our pipeline is composed of data preprocessing and machine learning techniques to formulate a prediction if a geomagnetic storm is to occur in the next 24 hours.

To the active sunspots and active sunspot regions counts extracted in the image processing layer for the present-day image, we also add the same features extracted from the previous day's image of the Sun. This helps numerically represent how drastically sunspots have changed on the Sun's surface just in one day. Additionally, we include a binary feature that tells us if a geomagnetic storm happened in the previous day. Adding this feature is extremely important because it provides input as to what the current atmospheric conditions are.

\subsubsection{Data Preprocessing}

To help the machine learning algorithm learn parameters much more efficiently, we apply a standardization algorithm for each element in the feature vector \(X_{i}\) on the \(i\)th day to create a standardized feature vector \(\hat{X_{i}}\):
\begin{equation}
    \hat{X_{i}} = \begin{bmatrix}
    \frac{X^{(i)}_{1}-min(X_{1})}{max(X_{1})-min(X_{1})}\\
    ...\\
    \frac{X^{(i)}_{5}-min(X_{5})}{max(X_{5})-min(X_{5})}
    \end{bmatrix}
\end{equation}

The standardized feature vector allows for the optimization process in the training stage of our machine learning algorithm to be much more efficient. Especially when using an SVM, feature scaling and standardization is almost a requirement \cite{noauthor_importance_nodate}.

\subsubsection{Gaussian Kernel SVM}

To forecast a geomagnetic storm, we use a \emph{Gaussian Kernel Support Vector Machine (G-SVM)} to formulate a prediction. An SVM is a supervised learning algorithm that is well-known for its ability to perform well in binary classification, as opposed to other supervised learning algorithms that are known for regression. The G-SVM is a variation of an SVM that creates a decision boundary in the data that is non-linear. A G-SVM we decide is the best choice of learning algorithm because of its ability to create a complex decision boundary for our high-dimensional data \cite{xue_nonlinear_2018}. From the training dataset, we train the G-SVM to predict if a geomagnetic storm is to occur in the next 24 hours; we feed to the G-SVM the 5 sunspots features described previously and the model will output \textit{storm} if the Kp-Index is predicted to be greater than or equal to 5, and \textit{no storm} otherwise.

\section{Evaluation}

In evaluating our techniques, we ask the following questions:

\begin{enumerate}
    \item Are the extracted sunspot features accurate with regards to the Internationally defined Space Environment Services Center Sunspot Number?
    \item How does the geomagnetic storm prediction test accuracy of our approach compare to NOAA's?
\end{enumerate}

\subsection{Experimental Setup and Datasets}
Our experiments were performed using OpenCV \cite{opencv_library} for computer vision and image processing techniques, and sci-kit learn \cite{scikit-learn} for machine learning techniques.

In selecting images of the Sun that would best show sunspot details, we determine that from NASA's Solar Dynamics Observatory (SDO), the HMI Flattened Intensitygram images of the Sun provided the most contrast between dark, active sunspots and light, inactive sunspots \cite{pesnell_solar_2012}. Images were then taken from the \(00:00:00\) hour of each day, so that the time the image was collected would correspond to the exact time that NOAA releases next-day predictions. In total, \(2843\) images were collected dating from January 2012 to April 2021.

To evaluate our feature extraction, we compare our results to the International Space Environment Services Center (SESC) Sunspot Number (the \emph{Wolf Number}), which is determined by the number of sunspot regions (\(r\)), the number of individual spots (\(s\)), and a vision constant (\(k\)) assigned to observers at the SESC to remove bias from approximations \cite{noauthor_sunspot_nodate}:
\begin{equation}
    \text{Wolf Number} = k(10r+s)
\end{equation}

Data on the SESC sunspot number was collected from the Sunspot Index and Long-term Solar Observations (SILSO) World Data Center \cite{noauthor_sunspot_nodate}.

Finally, to evaluate Kp-Index predictions from NOAA, we retrieved 1-day Kp-Index cycle predictions from the SWPC at NOAA. Then, for the comparison evaluation of our prediction, with NOAA's, we took the daily Kp-Index measurement data from the Helmholtz Centre Potsdam - GFZ German Research Centre for Geosciences \cite{matzka_geomagnetic_2021}. 

\subsection{Feature Extraction Accuracy}
To evaluate our features extraction layer, we compute the \emph{Pearson Correlation Coefficient} (PCC) between the features extracted from the Image Processing Layer, and the SESC Sunspot Number. The PCC is a statistical measure that finds the linear relationship between two random variables \cite{benesty2009pearson}. Since our algorithm does not include a vision constant \(k\) as defined in the SESC Sunspot Number, we use the PCC to quantify how similar or correlated our determined region and sunspot numbers is without multiplying their sum by an unknown \(k\) value. The PCC between two random variables \(X\) and \(Y\) is defined by their sample means, \(\bar{x}\) and \(\bar{y}\), and their respective i-th samples \(x_i\) and \(y_i\):
\begin{equation}
    \text{PCC} = \frac{\sum_{i=1}^{n} (x_i - \bar{x})(y_i - \bar{y})}{\sqrt{\sum_{i=1}^{n} (x_i - \bar{x})^2 \sum_{i=1}^{n} (y_i - \bar{y})^2}}
\end{equation}

From the population of region and sunspot numbers extracted from the Image Processing Layer, we create a new population \(X\) from region numbers \(R\) and sunspot numbers \(S\): 
\begin{equation}
    X = 10R + S
\end{equation}

\begin{figure}[t]
\centering
\begin{minipage}{.5\textwidth}
  \centering
  \includegraphics[width=\linewidth]{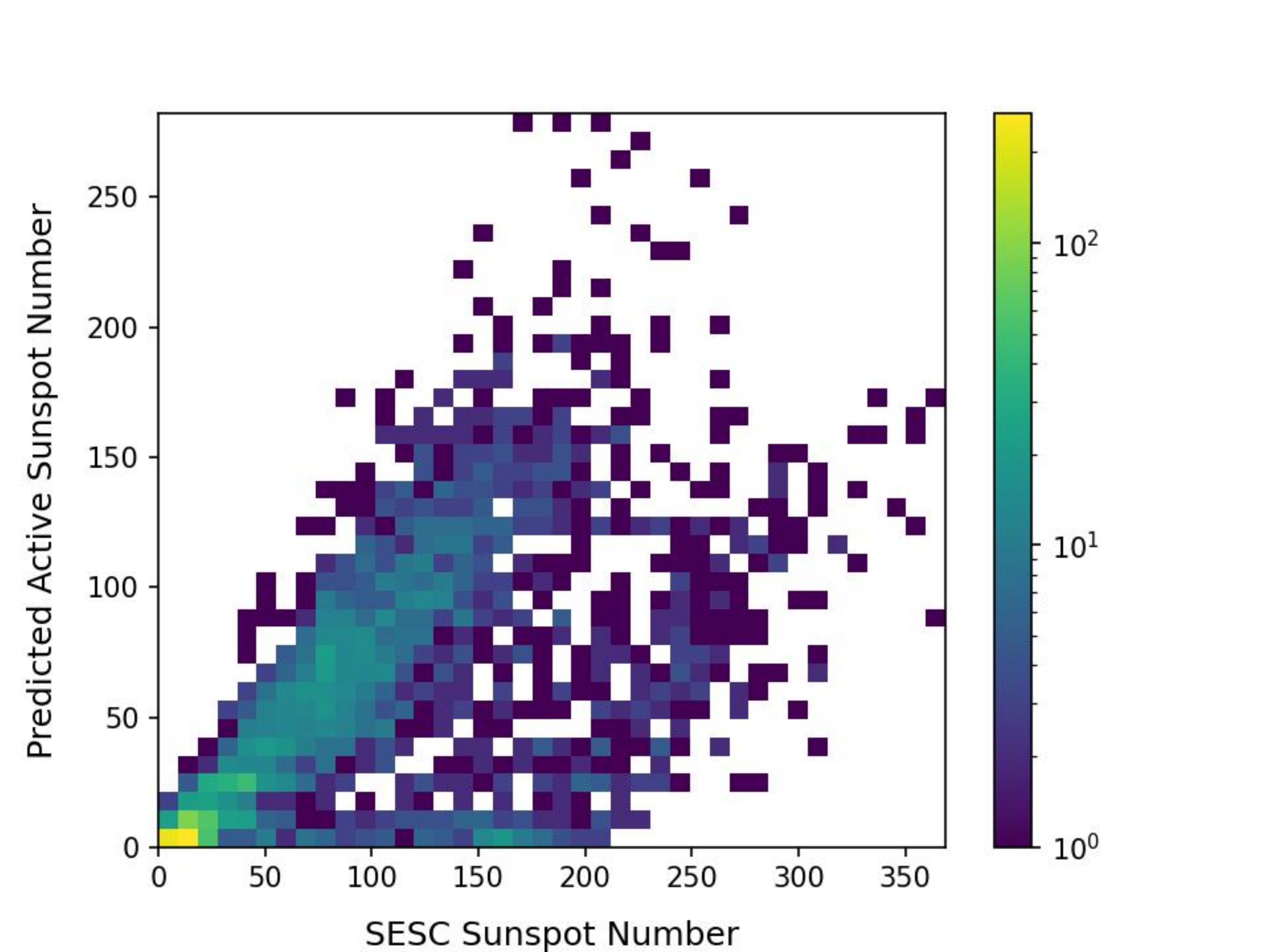}
  \captionof{figure}{Feature Correlation Density}
  \label{fig:pcc}
\end{minipage}%
\begin{minipage}{.5\textwidth}
  \centering
  \includegraphics[width=\linewidth]{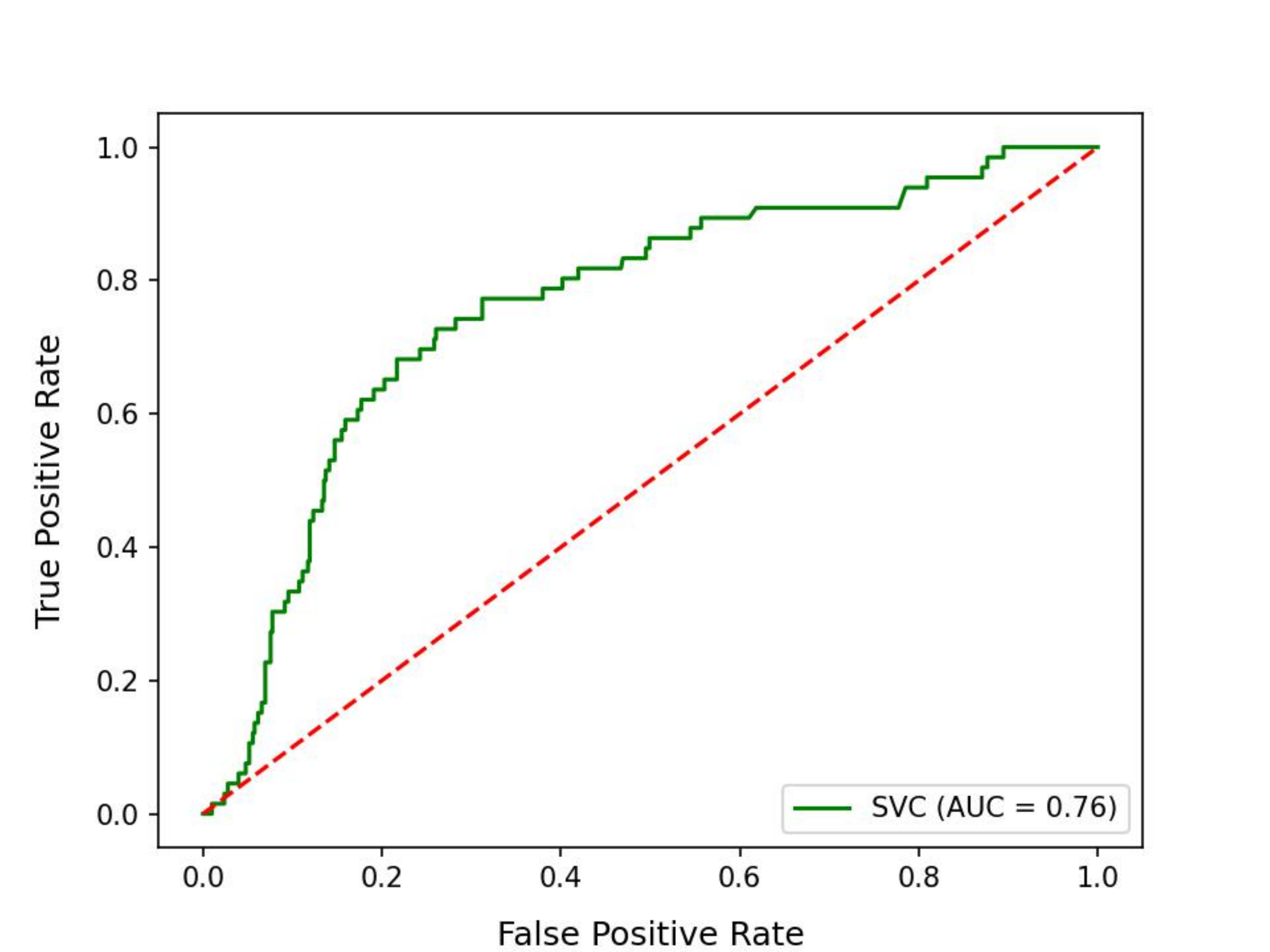}
  \captionof{figure}{ROC Curve on Predictions}
  \label{fig:roc}
\end{minipage}
\end{figure}

The PCC's domain is \([-1,1]\), where \(-1\) represents a 100\% negative correlation, and \(1\) a 100\% positive correlation. We then compute the PCC between the random variable \(X\) defined from our sunspot features, and a random variable \(Y\) representing the population of SESC sunspot numbers. A PCC of 0.66 was obtained, showing moderate to strong linear relationship between our features and the SESC Sunspot Number. 

The SESC Sunspot Number counts the total number of sunspot and sunspot regions (active and inactive). However, we specifically use hyperparameters in Canny that segment the darker, active sunspots and sunspot regions from the lighter, inactive ones. Evidence of this is shown when computing average loss between X and Y, which is \(-35\). Thus, explaining why a PCC of 0.66, in this case, is a very strong result; our Image Processing Layer was able to extract sunspot and sunspot region numbers from images of the Sun very efficiently.

\subsection{Prediction Layer Accuracy}

Using the features extracted in the Image Processing Layer, we test our Prediction Layer against NOAA's SWPC predictions with respect to the defined Kp-Index values. 

\subsubsection{Supersampling Techniques and Model Training }

From the \(2843\) data points from our dataset, 88\% of the data was composed of the \textit{no storm} geomagnetic storm class (Kp-Index value less than \(4\)). When supervised learning algorithms, such as an SVM, are trained on imbalanced data, the machine learning algorithm learns to only predict the majority class. To combat this severe class imbalance, we apply the \emph{Synthetic Minority Oversampling Technique} (SMOTE) algorithm to generate synthetic data points of the minority \textit{storm} geomagnetic storm class. SMOTE is the de facto algorithm for machine learning with imbalanced data, as it can effectively generate synthetic data in space where a minority class occupies \cite{fernandez_smote_2018}.

From the authentic dataset, we do an 80\% train-test-split, stratified by the minority class. Stratification allows us to balance the number of minority samples distributed among the train and test sets. From the train set, we perform synthetic oversampling with SMOTE, and train our G-SVM on the authentic data reserved for training, as well as synthetic data.

\subsubsection{Model Testing and Comparison}

From the \(2843\) original data points, a randomly selected 20\% of the data is reserved for testing. Stratifying the minority \textit{storm} classification, the test set was composed of \(503\) random \textit{no storm} classifications and \(66\) random \textit{storm} classifications. To evaluate the accuracy of the trained G-SVM, we plot a Receiver Operating Characteristic (ROC) curve on testing data. The ROC curve shows the efficacy of the decision boundary created by the G-SVM by plotting the false-positive rate over the true-positive rate in classification. The closer the curve is to the top left of the graph (\autoref{fig:roc}), the more accurate the G-SVM is considered. From the graph, Area Under the Curve (AUC) is used as the true accuracy of the classifier. An AUC value of 0.76 indicates that our model achieves a 76\% overall weighted accuracy across both classifications.

\begin{table}[!ht]
\centering
\begin{tabular}{|c|c|c|c|c|}
\hline
\textbf{Prediction Method} & \textbf{Class} & \textbf{Precision} & \textbf{Recall} & \textbf{Accuracy} \\ \hline
\multirow{2}{*}{G-SVM} & \textit{no storm} & 0.95 & 0.73          & \multirow{2}{*}{0.76} \\ \cline{2-4}
                       & \textit{storm}    & 0.26 & \textbf{0.73} &                       \\ \hline
\multirow{2}{*}{SWPC}  & \textit{no storm} & 0.94 & 0.90          & \multirow{2}{*}{0.86} \\ \cline{2-4}
                       & \textit{storm}    & 0.46 & \textbf{0.61} &                       \\ \hline
\end{tabular}
\caption{Classification Metrics Across Prediction Methods}
\label{tab:scores}
\end{table}

To compare our results to the SWPC at NOAA, we consider the 1-day storm prediction data provided from the SWPC. From the testing data used in the ROC curve, we retrieve the SWPC predictions for those same days, and evaluate based on precision, recall, and weighted accuracy scores on both methods. From \autoref{tab:scores}, precision and recall scores show that our model is competitive with the state-of-the-art industry prediction method with only using features collected from image processing, as opposed to collecting physical measurements from ground-based magnetometers and solar wind sensors from spacecraft in orbit.

\section{Conclusion}

This paper proposes a new approach to forecasting geomagnetic storms. With our solar system approaching another sunspot maximum, methods to predict such storms are becoming extremely important. Current prediction methods are limited in that they rely on solar wind measurements from spacecraft and magnetometer measurements from ground-based stations across the world. In this paper, we introduce a prediction method operating on sunspot features extracted by computer vision from images of the Sun. We show that machine learning techniques can leverage these sunspot features to accurately predict if a storm is to occur in the next 24 hours. Our algorithm consists of an image processing layer in which active sunspot features are collected via edge detectors and topological analysis. Then, active sunspot features are processed and used to forecast a geomagnetic storm with supervised learning techniques. Test accuracy is demonstrated to be competitive with the state-of-the-art model, indicating that sunspot features can be leveraged in concert with machine learning techniques to accurately forecast geomagnetic storms.

\bibliographystyle{splncs04}
\bibliography{mybib}

\end{document}